# Diffusion Deepfake

Chaitali Bhattacharyya[1]⋆, Hanxiao Wang[2]∗, Feng Zhang[3], Sungho Kim[1], and Xiatian Zhu[2]

[1] Yeungnam University, South Korea
[2] University of Surrey, UK
Nanjing University of Posts and Telecommunications, China

[https://surrey-uplab.github.io/research/diffusion_deepfake/](https://surrey-uplab.github.io/research/diffusion_deepfake/)

**Abstract.** Recent progress in generative AI, primarily through diffusion models, presents significant challenges for real-world deepfake detection. The increased realism in image details, diverse content, and widespread accessibility to the general public complicates the identification of these sophisticated deepfakes. Acknowledging the urgency to address the vulnerability of current deepfake detectors to this evolving threat, our paper introduces two extensive deepfake datasets generated by state-of-the-art diffusion models as other datasets are less diverse and low in quality. Our extensive experiments also showed that our dataset is more challenging compared to the other face deepfake datasets. Our strategic dataset creation not only challenge the deepfake detectors but also sets a new benchmark for more evaluation. Our comprehensive evaluation reveals the struggle of existing detection methods, often optimized for specific image domains and manipulations, to effectively adapt to the intricate nature of diffusion deepfakes, limiting their practical utility. To address this critical issue, we investigate the impact of enhancing training data diversity on representative detection methods. This involves expanding the diversity of both manipulation techniques and image domains. Our findings underscore that increasing training data diversity results in improved generalizability. Moreover, we propose a novel momentum difficulty boosting strategy to tackle the additional challenge posed by training data heterogeneity. This strategy dynamically assigns appropriate sample weights based on learning difficulty, enhancing the model's adaptability to both easy and challenging samples. Extensive experiments on both existing and newly proposed benchmarks demonstrate that our model optimization approach surpasses prior alternatives significantly.

---

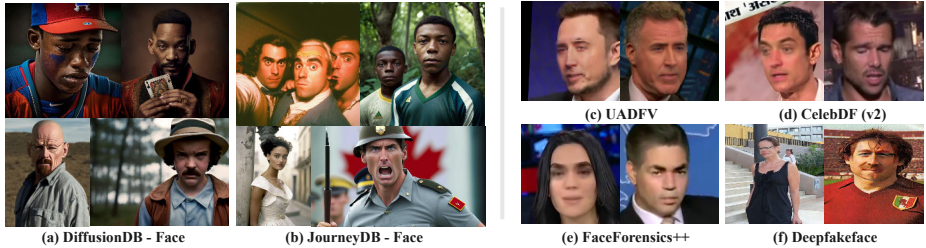

(a) DiffusionDB - Face  (b) JourneyDB - Face  (c) UADFV  (d) CelebDF (v2)
(e) FaceForensics++  (f) Deepfakeface

Fig. 1: Our proposed diffusion deepfake datasets **(a-b)** are featured with more realistic and faithful facial details and diverse background contents compared to the previous **(c-f)**.

## 1 Introduction

As more aspects of human life move into the digital realm, advancements in deepfake technology, particularly in generative AI like diffusion models [56], have produced highly realistic images, especially faces, which are almost indistinguishable to untrained human eyes. The misuse of deepfake technology poses increasing risks, including misinformation, political manipulation, privacy breaches, fraud, and cyber threats [29].

Diffusion-based deepfakes differ significantly from earlier techniques in three main aspects. Firstly, they exhibit **high-quality** by generating face images with realistic details, eliminating defects like edge or smear effects, and correcting abnormal biometric features such as asymmetric eyes/ears. Secondly, diffusion models showcase **diversity** in their outputs, creating face images across various contexts and domains due to extensive training on large datasets like LAION-5B, containing billions of real-world photos from diverse online sources [42]. Lastly, the **accessibility** of diffusion-based deepfakes extends to users with varying skill levels, transforming the creation process from a highly skilled task to an easy procedure. Even amateurs can produce convincing forgeries by generative models e.g., Stablility Diffusion [3] and MidJourney [2].

The rapid progress in deepfake creation technologies, fueled by diffusion models, has outpaced deepfake detection research in adapting to emerging challenges. Firstly, the lack of dedicated deepfake datasets for state-of-the-art diffusion models is evident. Widely used datasets like FF++ [40] and CelebDF [27] were assembled years ago using outdated facial manipulation techniques. The absence of a standardized diffusion-based benchmark impedes comprehensive assessment of deepfake detection models.

Secondly, existing research on deepfake detection often neglects the crucial issue of generalization. Many studies operate in controlled environments, training models on specific domains and manipulations and subsequently testing them on images from the same source. However, this approach falters when confronted with diffusion-generated deepfake images that span diverse domains and contents. Recent studies [55,11] highlight the struggle of deepfake detectors to gen-eralize to unseen manipulations or unfamiliar domains. Attempts to tackle this challenge, such as domain adaptation or transfer learning [6], have yielded suboptimal performance.

To address the identified problems, this paper presents two new deepfake detection benchmarks that utilize advanced diffusion models, namely ***DiffusionDB-Face*** and ***JourneyDB-Face***. These benchmarks encompass a wide range of content, incorporating diverse elements like head poses, facial attributes, photo styles, and realistic appearances. We expect these datasets to stimulate advancements in the identification of deepfakes generated through diffusion techniques. Our thorough assessment of these benchmarks indicates that the majority of current deepfake detectors, trained in constrained conditions, struggle to adapt to the evolving array of visual content generation methods, exemplified by diffusion models.

To enhance generalized deepfake detection, we advocate expanding the training data in terms of both scale and diversity. This approach is inspired by [36,33] that underscores the effectiveness of employing simple objective functions on extensive and diverse image datasets to achieve robust visual representations. In our initial pursuit of generalized deepfake detection, we suggest training a detector on an inclusive dataset covering a broad spectrum of deepfake generation techniques and image domains.

Acknowledging the varying complexities associated with different types of deepfakes, ranging from basic graphics-based face swaps to more intricate samples generated by diffusion models, we propose a novel momentum difficulty boosting strategy. This involves dynamically assigning different weights to samples based on their difficulties, thereby facilitating the model's adaptability to both straightforward and challenging deepfake samples.

This work contributes: (1) **Novel benchmarks**: We introduce two large-scale benchmarks, namely ***DiffusionDB-Face*** and ***JourneyDB-Face***, for deepfake detection. These benchmarks, designed to align with the rapid progress in generative AI models, offer a substantially increased number of high-quality face images with more diversity of images along with additional text description metadata. This surpasses the capabilities of previous benchmarks, creating notable challenges for existing detection models. Table 2 summarises the comparison between the conventional datasets and our proposed dataset. (2) **Generalizability assessment**: We extensively evaluate the generalizability of existing deepfake detection models on our new benchmarks. Operating under a challenging cross-domain scenario, our analysis uncovers the undesirable sensitivity of current models to domain shifts. This sensitivity often leads to a significant decline in performance. (3) **A novel generic training strategy for generation heterogeneity**: We show that our *momentum difficulty boosting* on datasets featuring diverse sources of deepfake generation methods markedly improves deepfake detection performance.

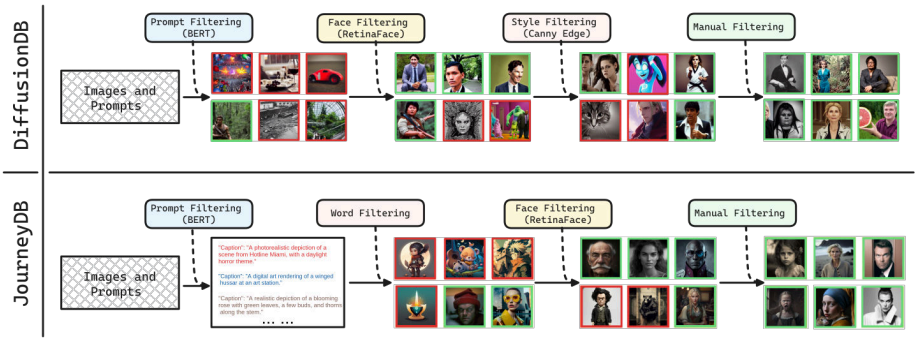

Fig. 2: Collection process for the proposed DiffusionDB-Face and JourneyDB-Face datasets. Green border : Images that were kept for the following round; Red Border : Images that were deleted after filtering.

## 2 Related Work

**DeepFake Creation and Benchmarks** The rise of deepfake technology poses a significant security threat, with the potential for misuse in spreading misinformation and engaging in malicious activities. In response, researchers are actively enhancing deepfake detection models to counter this threat. To evaluate these models, various datasets with diverse deepfake and authentic data from multiple sources have been established.

Earlier prominent deepfake datasets, such as FaceForensics++ [40], UADFV [57] and CelebDF [27], have been instrumental in this endeavor. The FaceForensics++ is created through four facial manipulation methods: FaceSwap [23], Face2Face [48], Deepfake [23] and NeuralTexture [47]. It also provides three compression levels to evaluate detectors under varying compression scenarios. UADFV creates fake face images by splicing face region synthesized using deep neural network into the original image. Nevertheless, these datasets exhibit low visual quality, markedly differing from Deepfake videos disseminated on the internet. Consequently, the CelebDF dataset focuses on achieving superior visual quality through an AutoEncoder-based deepfake synthesis method, including 590 real videos and 5639 synthetic celebrity videos.

With the advancement of generative models, there has been a proliferation of highly realistic Deepfake videos produced by a multitude of GAN variants [14,35]. However, GAN-based deepfake methods still face limitations, notably the absence of realistic backgrounds in the generated images [8,30,53].

Diffusion models [39] have gained widespread attention due to their ability to generate visually plausible content. Ricker *et al.* [38] demonstrated through extensive evaluation experiments that identifying images generated by diffusion models is a more challenging task than recognizing GAN-generated images. In contrast to GANs, deepfake images generated by diffusion models do not exhibit noticeable grid-like artifacts in the frequency domain. Song *et al.* [45] uti-lized diffusion models to create a synthetic celebrity face dataset, Deepfakeface. They similarly introduced two new tasks to enhance the assessment of detection methods performance. In parallel, we propose two deepfake datasets based on diffusion models: DiffusionDB and JourneyDB. Compared to Deepfakeface, our benchmarks cover a wider range of content and importantly exhibit more significant challenges to existing deekfake detectors (see Tables 4 and 6). Table 2 summarises three conventional and three diffusion generated benchmark datasets (including our dataset). The table shows that our dataset is bigger than other datasets with more diversity per images and also contains metadata. Supplementary material has more samples from the dataset proving the *diversity* in our dataset which other dataset lacks. By incorporating cutting-edge diffusion models, the deepfake images in these datasets feature diverse elements like head poses, facial features, and image styles while exhibiting a realistic appearance. We expect these datasets to drive progress in detecting deepfake generated by diffusion models.

**DeepFake Detection** relies on analyzing different feature signals to ascertain the authenticity of an image. Earlier efforts focused on analyzing physiological signals for deepfake detection. Li *et al*. [25] identified the absence of eye blinking as a telltale sign for detecting deepfake videos and showed that distinguishing open and closed eye states could help. Additional efforts have explored features such as head poses [57], speaking-action patterns [4], and the combinations of various physiological signals [9].

Furthermore, many methods involving the search for potential synthetic artifacts and analysis of local features have been proposed. FWA [26] detects deepfakes by simulating facewarping artifacts. Face X-ray [24] predicts the presence of blending boundaries. Zhu *et al*. [60] introduced 3D decomposition into deepfake detection, amplifying subtle local artifacts through facial detail construction and detection. Recent research like DIRE [50] use image reconstruction error as a differentiating factor between real and fake images for detection. Frequency domain cues are also crucial for distinguishing deepfakes. Luo *et al*. [28] highlighted that CNN-based detectors tend to overfit to color textures in cross-database scenarios, suggesting the use of high-frequency noise for face forgery detection.

Data-driven approaches aim to directly learn how to differentiate real images from deepfakes through various strategies, exhibiting better generalization [31,49,20,15,44]. Capsule [31] pioneers the use of capsule networks in the deepfake detection task. Wang *et al*. [49] emphasized the importance of careful pre- and post-processing and data augmentation to enhance the generalization. Recently, Guo *et al*. [20] proposed a hierarchical fine-grained formulation to address the diversity of images generated by various forgery methods. By encouraging the model to learn integrated features and inherent hierarchical properties of different forgery attributes, this approach improves deepfake detection representation.

In this work, we emphasize the importance of using heterogeneous training images for extended model generalization. Further, a novel model-agnostic mo-mentum difficulty boosting strategy is introduced for more effective training by dynamically tuning the weights of individual samples during optimization.

## 3 Diffusion DeepFake Benchmarks

AI-generated content (AIGC) platforms like DALL-E, Stability AI, and Midjourney empower global users to craft detailed, high-quality images from text prompts. Several general large-scale prompt-to-image datasets, e.g. the MidjourneyDB [34] and DiffusionDB [51], have thus been collected by crawling from public sources (e.g. Stable Diffusion and Midjourney Discord servers). Our approach to constructing diffusion-based deepfake datasets involves iterative textual and visual filtering of these general prompt-image datasets. This curation process aims to refine prompts/images progressively, ensuring they exclusively feature high-quality human face images.

### 3.1 DiffusionDB-Face Construction

We initiated our dataset curation with the DiffusionDB(2M) dataset [51], comprising 2 million images generated by Stable Diffusion, each associated with prompts. To curate a deepfake dataset which only contains high-quality face images, we design an iterative approach with four steps of filtering following a coarse-to-fine progression:

(1) **Prompt Filtering by LLM**:

The goal of this step is to quickly reduce the candidate prompt pool such that only prompts related to human faces are retrieved. Inspired by the outstanding zero-shot capability of large-language models (LLM), we defined a zero-shot classification task to classify the associated prompt of each image into two predefined categories ("human face", "not human face") with the HuggingFace Transformer toolbox [52]. We used a pre-trained language model (BERT-base [13] with 12 transformer blocks, 12 attention heads, 110M parameters) to classify each prompt in the original DiffusionDB to obtain a prediction score for the pre-defined class of "human face" (see Figure 4). We set a threshold value of 0.5 and discarded all the prompts whose prediction were smaller than the thershold. With this approach we successfully removed $95 + \%$ of the original prompts not semantically related to human faces.

(2) **Detection based auto filtering**: With 84,830 prompts remaining after the first step, we employed the state-of-the-art RetinaFace detector [12] on all associated images to selectively retain those featuring human faces. In this phase, we utilized the RetinaFace model with its default configuration, setting the confidence threshold at 0.5. Images where the confidence score meets or exceeds this threshold are retained for subsequent filtering stages. Utilizing RetinaFace, we successfully extracted most images containing faces (see Figure 5b).

Having obtained 39,887 human face images, a quick manual inspection revealed various unrealistic images with distinct artistic styles (e.g., black-and-white / anime / cartoon / sketch-style faces).

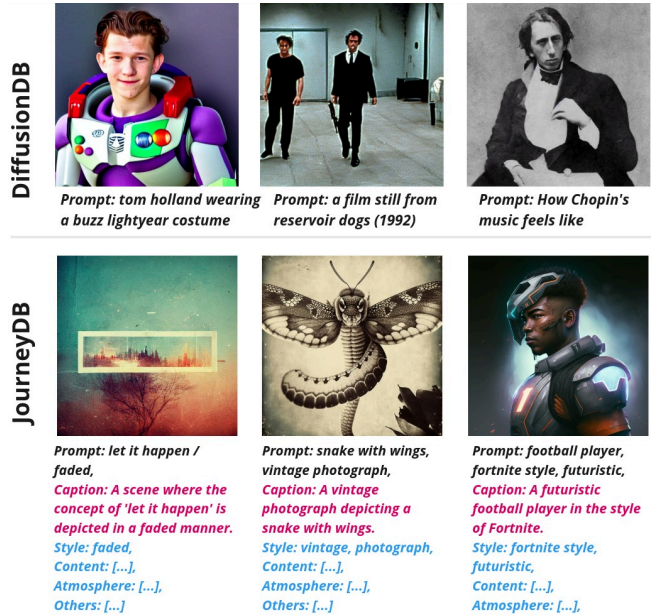

Fig. 3: Input metadata along with the corresponding images.

(3) **Edge/color based style filtering**: We adopted two additional steps to further refine our data. (I) We measure the color variance of the original image to identify whether the input image has a too narrow color spectrum; (II) We apply a Canny edge detector to measure the number of edges on the images to identify images with specific drawing styles and animations. Empirically we set the edge threshold at 100 and the color threshold at 200 to determine whether an image is with unrealistic style, and excluded images if their edges exceeded the edge threshold or color variance falls below the color threshold. This step helped to reduce about 50% images from the last round.

(4) **Manual filtering**: In the final step, we conducted a manual annotation process, resulting in a curated dataset of 18,371 high quality realistic human faces, which we refer as DiffusionDB-Face.

### 3.2 JourneyDB-Face Construction

To retrieve face images from JourneyDB [34] suitable for deepfake detection, we followed the same procedure as in Sec 3.1, with three minor adjustments. (1) To ensure we have enough test images in our deepfake detection benchmark, we ignored the original train / validation / test split provided by JourneyDB.

(2) Since the metadata of JourneyDB also include style prompts (see Figure 3), we thus replaced the edge/color-based style filtering step as in DiffusionDB to an exclusive word filtering on the style prompts to remove images with unrealistic styles such as "Anime Style", e.g. see Figure 5a.

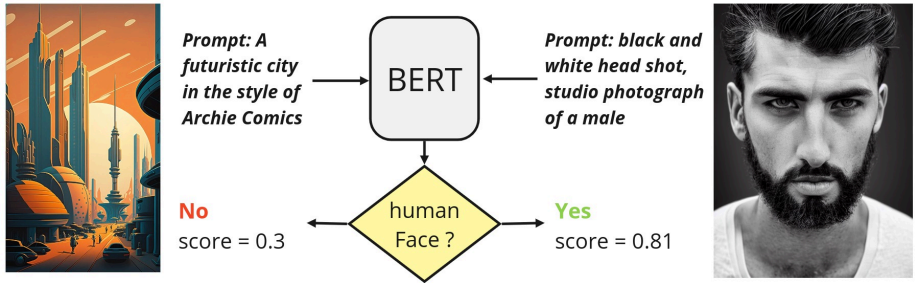

Fig. 4: Example of prompt filtering by language model. Note, only text is the input to BERT [13], whilst the associated image is shown for illustration only.

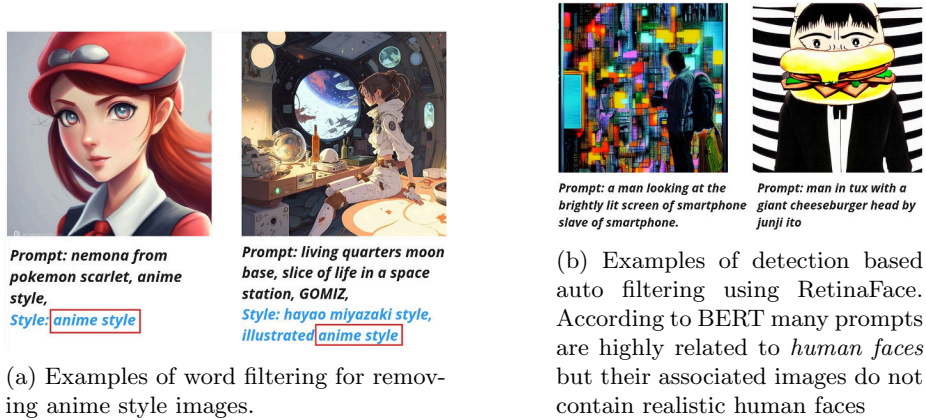

(a) Examples of word filtering for removing anime style images.

(b) Examples of detection based auto filtering using RetinaFace. According to BERT many prompts are highly related to *human faces* but their associated images do not contain realistic human faces

Fig. 5: Illustrations of prompt filtering.

(3) The test partition of the JourneyDB dataset does not come with any metadata, so we direcly applied RetinaFace detector followed by a manual filtering process.

### 3.3 Data Preprocessing

The basic statistics of DiffusionDB-Face and Journey-Face are shown in Figure 7. We used the Deepface [43] framework to analyze the gender distribution statistics within our dataset. After the acquisition of the datasets, a comprehensive preprocessing pipeline was executed to optimize the data for utilization in deep learning architectures and to facilitate ease of analytical operations. Specifically, we performed a re-examination of each image for facial detection by MTCNN [58]. Some images were further discarded at this stage due to face detection failures or only containing too small faces without enough visual details for the deepfake detection task.

After face detection, the images were uniformly cropped to a resolution of $256 \times 256$ pixels, establishing a standard input size.

Finally, the preprocessed dataset with standardized face detection crops has 24,794 and 87,833 deepfake images for propsoed DiffusionDB-Face (DFDB-Face) and JourneyDB-Face (JDB-Face) benchmark respectively. Subsequently, these images were categorized into train / test / val subsets with a 90 : 5 : 5 ratio respectively as shown in Table 3.

Table 1: Number of images after each round of processing: DiffusionDB and JourneyDB

| Dataset | INPUT | Round 1 | Round 2 | Round 3 | Round 4 | Preprocessed (Final) |
|---|---|---|---|---|---|---|
| DiffusionDB-Face | 2,000,000 | 84,830 | 39,887 | 18,845 | 15,198 | 24,794 |
| JourneyDB-Face | 4,932,309 | 238,869 | 225,759 | 78,904 | 61,984 | 87,833 |

Table 2: Dataset summary. Top: Conventional datasets; Bottom: Diffusion datasets; V: Video datasets. MF/S : Multiple faces per sample ; Generation: Generation methods.

| Source | No. Fake | No. Real | Generation | Metadata | MF/S |
|---|---|---|---|---|---|
| FF++ [40] (V) | 4,000 | 977 | F2F [48],DF [23],FS [23],NT [47] | ✗ | ✗ |
| UADFV [57] (V) | 49 | 49 | FS [23], DF [23] | ✗ | ✗ |
| CelebDFv2 [27] (V) | 5,639 | 590 | Autoencoder [27] | ✗ | ✗ |
| DeepFakeFace [45] | 3×30,000 | 30,000 | SD [3], IP [3], IF [1] | ✗ | ✓ |
| **JDB-Face (ours)** | **87,833** | **94,120** | Midjourney [2] | ✓ | ✓ |
| **DFDB-Face (ours)** | **24,794** | **94,120** | SD [3] | ✓ | ✓ |

Table 3: Data split per dataset.

| Dataset | Train | | Test | | Validation | |
|---|---|---|---|---|---|---|
| | Real | Fake | Real | Fake | Real | Fake |
| CelebDF V2 [27] | 35,469 | 160,595 | 1,971 | 8,922 | 1,971 | 8,922 |
| FF ++ [40] | 17,847 | 102,755 | 990 | 5,711 | 993 | 5,711 |
| UADFV [57] | 1,393 | 1,371 | 77 | 78 | 76 | 77 |
| Deepfakeface [45] | 27,000 | 81,00 | 1,500 | 4,500 | 1,500 | 4,500 |
| JDB-Face (ours) | 82,440 | 78,757 | 4,581 | 4,375 | 4,580 | 4,376 |
| DFDB-Face (ours) | 82,440 | 22,331 | 4,581 | 1,241 | 4,580 | 1,241 |

Additionally, to evaluate the full classification performance, we have sourced 94,120 authentic face images from the Flickr-Faces-HQ (FFHQ) dataset [22] so that we can measure both the sensitivity and specificity of the deepfake detection methods.

## 4 Momentum Difficulty Boosting

The conventional deepfake detection training and evaluation protocol tends to overlook the critical issue of generalization, often yielding inflated detection performance. Specifically, a detector may exhibit impressive results when trained and tested on deepfakes generated from the same source, within a limited range of manipulations and image domains. However, as observed in [55,10] and corroborated by our subsequent evaluations, these detectors experience a substantial performance drop when applied to deepfakes from different sources/domains. This challenge is particularly pronounced, as demonstrated in Sec 5.1, when detectors are applied to the diverse diffusion-generated deepfakes. To address this limitation, we advocate for a new setting, where the performance of a detector should be benchmarked against multi-source training and test datasets, providing a more comprehensive understanding of its generalizability across various domains.

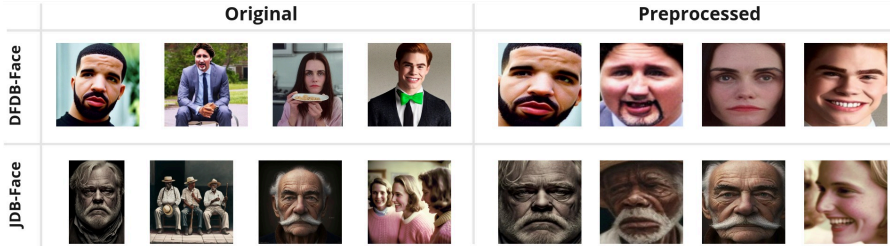

Fig. 6: Visualization before and after preprocessing the images.

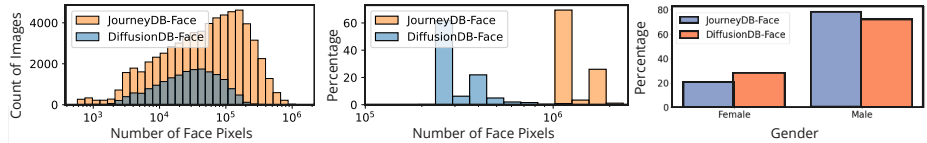

Fig. 7: Basic statistics of our datasets.

We begin with a set of $K$ diverse deepfake datasets $\{\mathcal{D}^1, \mathcal{D}^2, \cdots, \mathcal{D}^K\}$. Each dataset $\mathcal{D}^k = \{(\mathbf{x}_i^k, y_i^k)\}_{i=1}^{N_k}, k \in [K]$, comprises $N_k$ images sourced from specific domains and deepfake manipulation methods. For instance, one dataset may include Instagram-style selfies with deepfakes generated using diffusion models. We further denote $f_\theta$ the target deepfake detection model parameterized by $\theta$, $\hat{y}_i^k = f_\theta(\mathbf{x}_i^k)$ the model prediction, and $\ell(y_i^k, f_\theta(\mathbf{x}_i^k))$ a general loss function in the context of deepfake detection, e.g. a standard binary cross-entropy loss or more advanced loss designs as in [32,54].

**Conventional Setting** Existing methods [38,19,20] often train deepfake detection models individually on each $\mathcal{D}^k$, and evaluate each trained model using the corresponding test set $\bar{\mathcal{D}}^k$, where images are sampled from the same source. Formally, they attempt to optimize the objective:

$$\min_{\theta_k} \mathbb{E}_{x_i^k \in \mathcal{D}^k} \left[ \ell(y_i^k, f_{\theta_k}(\mathbf{x}_i^k)) \right] + \lambda \mathbf{R}(\theta_k), \tag{1}$$

where the first and second term correspond to the empirical loss on $\mathcal{D}^k$ and the regularization term, respectively. However, this approach makes the unrealistic assumption that the image domains and manipulation methods are known during deployment, suffering from significant performance drop when facing domain and forgery type shifts.

**Proposed Setting** Instead of employing domain and manipulation-specific models, our objective is to train a single model $f_\theta$ agnostic to data source $k$. We combine images from $\{\mathcal{D}^k\}_{k=1}^K$ into a heterogeneous dataset denoted as $\mathcal{D}_H = \{(\mathbf{x}_i, y_i, k_i)\}_{i=1}^{\sum_k |\mathcal{D}_k|}$.

As shown in our later experiments in Sec 5.1, directly training on such a mixed dataset did not translate to good cross-dataset performance, due to additional challenges imposed by diverse training samples with various level of difficulty.

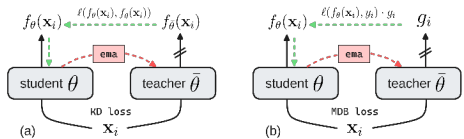

Fig. 8: (a) Momentum-based knowledge distillation [7]; (b) Our MDS: only use the 'teacher' network to weight samples by their difficulties.

**Momentum Difficulty Boosting** We thus propose to employ a boosting function to ease the training with data heterogeneity. This function regulates the importance of examples, assigning more weights to the difficult ones. Specifically, $g_i = g(\mathbf{x}_i, y_i, \theta)$ quantifies the instantaneous instance difficulty of sample $\mathbf{x}_i$, considering the under-optimized model parameters $\theta$. Our revised optimization objective thus becomes

$$\min_\theta \mathbb{E}_{\mathbf{x}_i \in \mathcal{D}_H} [g_i \times \ell(y_i, f_\theta(\mathbf{x}_i)] + \lambda \mathbf{R}(\theta). \tag{2}$$

We proposed a simple yet effective strategy, momentum difficulty boosting (MDB), to calculate the sample-wise difficulty scores. Specifically, we maintain a momentum moving-average of the detector, $\bar{\theta}$, and use it to calculate sample difficulties on-the-fly by measuring the cross-entropy between the momentum network's prediction and the data samples' ground truths. Formally, we define the sample-wise difficulty score as

$$g(\mathbf{x}_i, y_i, \theta) = CE(y_i, f_{\bar{\theta}}(\mathbf{x}_i)), \tag{3}$$

where $\bar{\theta}$ slowly tracks the detector's parameter $\theta$ by the momentum updating rule: $\bar{\theta} = m\bar{\theta} + (1-m)\theta$.

The momentum update's benefit lies in mitigating the substantial variance in predicted sample difficulty scores, thereby enhancing training stability. Our approach shares conceptual similarities with knowledge distillation [21,7,5], with the difference that instead of directly distilling knowledge from a teacher network $\bar{\theta}$, we leverage it as a guiding function to adjust the training data distribution by assigning different weight to each sample based on their difficulty levels. During training, the sample weights $g_i$ are decided by $\bar{\theta}$ based on in Eq. (3), where both the weights and $\bar{\theta}$ are updated dynamically at each mini-batch (see Figure 8). To prevent domination of certain samples with exceptionally high difficulty scores, we re-scale the sample weights in each mini-batch to fall within the range $[1, C]$, where $C$ denotes the capped maximum sample weight.

## 5 Experiments

### 5.1 Evaluation of off-the-shelf models

We first produce a comprehensive evaluation of a range of existing pre-trained deepfake detectors on the generalization capability to understand how their performance degrade when tested on deepfake images from different sources/domains than training, especially on the newly collected diffusion-based deepfakes from our DiffusionDB-Face and JourneyDB-Face datasets.

**Datasets** We consider three conventional datasets and three diffusion-based datasets in our evaluation also summarized in Table 2. (1) *FaceForensics++ (FF++)* [40] consists of 1,000 video clips designed for digital forensics. It encompasses four facial modification techniques, including Face2Face [48], Deepfakes[23], FaceSwap[23], and NeuralTextures[47]. This dataset contains 977 YouTube videos, each featuring front-facing, easily trackable faces. (2) *CelebDFv2* [27], includes genuine YouTube videos and synthesized deepfake videos. In its first version, there are 408 genuine videos and 795 deepfake videos, covering diverse characteristics like ethnicity, age, and gender. The second version extends the dataset with 590 genuine videos and 5,639 deepfake videos obtained from online sources, further increasing data diversity. (3) *UADFV* [25] includes 49 genuine videos collected from the internet and then manipulated by [23] to generate deepfakes. (4) *Deepfakeface* [45] includes 90,000 fake images from three different generation methods i.e. StableDiffusionv1.5 [3], Inpainting [3] and InsightFace [1], along with 30,000 real images. (5-6) Our *DiffusionDB-Face* and *JourneyDB-Face* include various diffusion-based deepfakes generated by two art generative AI providers, Stability AI and MidJourney. The dataset collection process are detailed in Sec 3.1 and 3.2. (7) Fake-CelebA [50] was formed using four diffusion generation methods (a) SD-v2 [39] (42,000 images), (b) IF [41] (1,000 images), (c) DALLE-2 [37] (500 images), (d) Midjourney [2] (100 images), along with 42,000 real images. Train/test/validaiton split of the datasets is summarised in Table 3.

**Competitors** We consider seven pre-trained deepfake detection models. Specifically, (1) *HiFi Net* [20] is a fine-grained deepfake detector based on multi-branch feature extraction and hierarchical forgery predictions, trained on a customised dataset with a taxonomy of image forgery types ranging from CNN-based manipulations to image editing. (2) *SBIs* [44] is trained on FF++ with a novel image blending method reproducing common forgery artifacts, e.g., blending boundaries and statistical inconsistencies. (3) *CADDM* [16] is trained on FF++ with a constraint to mitigate the effect of identity leakage whilst performing deepfake detection. We used its EfficientNet-b4 variant in our evaluation. (4) *CNNDet* [49] trains a ResNet50 model on a customized dataset of deepfakes solely generated by ProGAN [17]. (5) *DSP-FWA* [26] is a deepfake detector specifically aiming to detect the warping artifacts of the deepfake creation process, trained with real images collected from Internet and a customized algorithm to generate negative data with warping effects. We used its SPP-Net varriant in our evaluation. (6) *Capsule* [31] is a Capsule network-based deepfake detector trained with the FF++ dataset. (7) *DIRE* [50] is a diffusion model generated deepfake detection method where a novel image representation is introduced to measure the error between input images.

**Setting** We followed the evaluation protocol proposed in [55]. For FF++ dataset, we have considered it as a unified dataset rather than separating it into four different parts with seperate manipulations. All the images from each dataset were preprocessed and cropped into a size of $256 \times 256$. The video datasets were sampled into frames, i.e. we took 32 frames per video after detecting the frames that included faces. Specifically, 19,830/114,213 (real/fake) video frames are sampled for FF++, 1,548/1524 for UADFV and 39,411/178,439 for CelebDFv2. All the listed deepfake detectors were evaluated with their officially released pretrained weights and directly applied to the test splits of each dataset without further finetuning. We adopt three metrics for evaluation, including AUC (area under the ROC curve), EER (equal error rate), and ACC (accuracy).

Table 4: Evaluation performance of off-the-shelf DeepFake detectors on conventional deepfake datasets (FF++, CelebDFv2, UADFV) and diffusion deepfake datasets (Deepfakeface, DFDB-Face, JDB-Face, Fake CelebA). Highest accuracy in **bold**.

(a) Conventional deepfake datasets (FF++, CelebDFv2, UADFV)

| Model | FF++ | | | CelebDFv2 | | | UADFV | | |
|---|---|---|---|---|---|---|---|---|---|
| Metric | AUC | EER | ACC | AUC | EER | ACC | AUC | EER | ACC |
| HiFi Net | 0.60 | 0.41 | **0.58** | 0.60 | 0.41 | **0.58** | 0.60 | 0.45 | 0.54 |
| SBIs | 0.58 | 0.43 | 0.56 | 0.51 | 0.72 | **0.67** | 0.51 | 0.74 | 0.50 |
| CADDM | 0.50 | 0.48 | 0.52 | 0.50 | 0.50 | 0.50 | 0.56 | 0.46 | **0.53** |
| CNNDet | 0.76 | 0.29 | **0.71** | 0.54 | 0.46 | 0.53 | 0.53 | 0.41 | 0.58 |
| DSP-FWA | 0.54 | 0.61 | 0.33 | 0.66 | 0.40 | 0.51 | 0.48 | 0.51 | 0.48 |
| Capsule | 0.80 | 0.26 | **0.73** | 0.61 | 0.43 | 0.56 | 0.79 | 0.29 | 0.71 |
| DIRE | 0.11 | 0.91 | 0.22 | 0.14 | 0.90 | 0.21 | 0.22 | 0.85 | 0.27 |

(b) Diffusion deepfake datasets (Deepfakeface, DFDB-Face, JDB-Face, Fake CelebA)

| Model | FF++ | | | CelebDFv2 | | | UADFV | | | Fake CelebA | | |
|---|---|---|---|---|---|---|---|---|---|---|---|---|
| Metric | AUC | EER | ACC | AUC | EER | ACC | AUC | EER | ACC | AUC | EER | ACC |
| HiFi Net | 0.57 | 0.45 | 0.45 | 0.52 | 0.66 | 0.51 | 0.45 | 0.68 | 0.40 | 0.51 | 0.55 | 0.49 |
| SBIs | 0.51 | 0.61 | 0.50 | 0.25 | 0.89 | 0.30 | 0.41 | 0.82 | 0.49 | 0.57 | 0.45 | 0.54 |
| CADDM | 0.51 | 0.49 | 0.50 | 0.48 | 0.70 | 0.47 | 0.52 | 0.73 | 0.52 | 0.51 | 0.68 | 0.48 |
| CNNDet | 0.61 | 0.41 | 0.58 | 0.53 | 0.49 | 0.52 | 0.44 | 0.75 | 0.45 | 0.40 | 0.62 | 0.58 |
| DSP-FWA | 0.50 | 0.88 | 0.40 | 0.52 | 0.51 | **0.54** | 0.52 | 0.47 | 0.53 | 0.38 | 0.57 | 0.42 |
| Capsule | 0.49 | 0.49 | 0.50 | 0.48 | 0.57 | 0.46 | 0.45 | 0.56 | 0.46 | 0.49 | 0.68 | 0.50 |
| DIRE | 0.38 | 0.76 | 0.55 | 0.62 | 0.45 | 0.71 | 0.42 | 0.54 | 0.51 | 0.68 | 0.34 | **0.72** |

**Results** As shown in Table 4, we have made the following observations:

(1) All pre-trained detectors exhibit pronounced generalization issues when tested on deepfakes originating from different sources or domains. For instance, the Capsule model [31], trained on the FF++ dataset, achieved a high AUC of 0.80 on the same dataset. However, its AUC dropped to 0.61 on CelebDFv2 generated by a different deepfake method. On the three diffusion-based deepfake datasets, its performance further degraded, with AUC decreasing to 0.49, 0.48, and 0.45 for Deepfakeface, DiffusionDB-Face, and JourneyDB-Face, respectively. On the other hand DIRE [50] has performed comparatively better with Fake CelebA but still not upto the mark due to the absence of the same domain's dataset in the training. This observation strongly highlights the generalization issue of existing deepfake detectors, impeding their practical utility in real-world scenarios where deepfakes can emerge from diverse sources and domains.

(2) Among all datasets, the diffusion-based ones have proven to be the most challenging for existing deepfake detectors. This is evident in the substantial performance gap between the three conventional datasets and the diffusion ones. Notably, on the proposed DiffusionDB-Face and JourneyDB-Face, all examined detectors (except DIRE) obtain AUC values below 0.55, indicating even worse performance than random guessing. However, even with DIRE detector, we JDB-Face performed worst among all diffusion datasets with 51% accuracy. This suggests that highly realistic facial images generated by the latest diffusion models can easily confuse pretrained deepfake detectors, leading them to be frequently misclassified as real faces and thus remaining undetected.

## 5.2 Evaluation Under Varying Evaluation Strategies

**Base detection model** We use the Capsule network [31] as the base deepfake detector including our MDB. The reason behind is due to its ability to achieve consistently top performance on most datasets.

**Competitors** We consider two training settings: (1) *Single-domain training:* The deepfake detector is trained using one dataset/domain. We repeat this practice for all six datasets with the base detector as specified in Sec 5.1. (2) *Multi-domain training:* We combine six datasets (FF++, CelebDFv2, UADFV, Deepfakeface, DFDB-Face, JDB-Face) with different deepfake types by simple concatenation and shuffling and train the deepfake detector. This training strategy includes the following methods: We compare the following training methods: *(a) Vanilla:* Training the base detector on the merged 6 datasets. *(b) Knowledge Distillation (KD):* [18] We replace the dynamic difficulty weighting process with a knowledge distillation loss [18] between $\bar{\theta}$ and $\theta$. This comparison aims to evaluate the proposed MDB strategy against a knowledge distillation approach, as both require a separate network $\bar{\theta}$ to be maintained during training. *(c) Difficulty Weighing (DW) [46]:* We use difficulty weighting without momentum i.e. we directly use the in-training network $\theta$ to generate the difficulty scores, without referring to the momentum-updated network $\bar{\theta}$. This comparison is intended to evaluate the effectiveness of the proposed MDB strategy. *(d) Our proposed MDB:* We set the momentum $m = 0.97$ and the sample weight rescale factor $C = 5$. For all training strategies, we trained *from scratch* with randomly initialized weights and used the same hyper-parameters with a learning rate of 0.0001, momentum for Adam optimization of 0.9 and the alpha value of 0.99.

**Cross-domain test** We further evaluate the models trained as above on an unseen domain. We choose the Fake-CelebA [50] as the test dataset for the multi-domain training setting. This dataset has been generated by four diffusion models.

**Results** From Tables 6, we observe that:

(1) *Single-domain training* on the diffusion deepfakes improved the detector's performance on this new deepfake type. Specifically, the Capsule model's evaluation accuracy was boosted to 0.68/0.73/0.67 from 0.50/0.39/0.39 on Deepfake-face, DiffusionDB-Face, and JourneyDB-Face, when we train it from scratch on each of these datasets. However, we also noticed that this improvement comes with a sacrifice on other datasets. For example, the JourneyDB-Face trained model achieved poor accuracies on all conventional deepfake datasets, with an average value of only 0.39.

(2) Directly training a model with *vanilla* method helped improve deepfake detection performance across all datasets, but only to a limited extent. Specifically, we observe an average accuracy across the six datasets of 0.43 with the multi-domain training, a small increase compared to the individual single-domain trainings (except for FF++).

(3) In comparison with standard *knowledge distillation* (referred as *KD* in Table 6), which achieves an average accuracy of 0.70, the proposed MDB exhibits a 20% improvement. Similar observations can be made when comparing it with the naive difficulty weighting strategy without momentum updating (referred as *DW* in Table 6), which has an average accuracy of 0.59. Such observations show that the proposed MDB's improvement is non-trivial. (4) Our proposed *MDB* led to a substantial performance gain with multi-domain training. By dynamically assigning sample weights according to their difficulties, it perfectly aligns with the diverse nature of the multi-domain training set and enables the model to focus on more difficult samples along the training. Specifically, we see an average accuracy of 0.76/0.92/0.84 for the proposed MDB approach on conventional/diffusion/all dataset respectively. The corresponding AUC and EER values show much higher ability to distinguish between real and fake images even with the unbalanced datasets. For example, our strategy has 0.94 (AUC) for FF++ (non-diffusion dataset) and 0.93(AUC) for JDB-Face (diffusion dataset). However, the results for UADFV is not up to the mark. This is due to a much smaller training set (1.3k fake images) with UADFV, in comparison to 102k for FF++, 160k for CelebDFv2, and 62k for JDB-Face.

Table 5: Performance metrics comparison

| Metric | ACC | EER | AUC |
|---|---|---|---|
| Vanilla | 0.57 | 0.58 | 0.49 |
| KD [18] | 0.67 | 0.60 | 0.44 |
| DW [46] | 0.54 | 0.60 | 0.50 |
| **MDB (ours)** | **0.80** | **0.21** | **0.78** |

(5) We use Fake-CelebA [50] as totally unseen data for cross-domain generalisation test. The results in Table 5 illustrate superior outcomes by our MDB when applied to a distinct or unfamiliar domain of diffusion-generated images. This validates the advantages of our proposed method compared to the other competitors. We have added more ablative analysis in *supplementary material*.

Table 6: Comparison of generalization capabilities across different datasets and training strategies using the Capsule network as the base deepfake detector. Accuracy (ACC), Equal Error Rate (EER), and Area Under the Curve (AUC) metrics are presented. The best results are in **bold**. The top part of each sub-tables shows the single-domain training setting.

(a) Conventional deepfake datasets (FF++, CelebDFv2, UADFV)

| Train Strategy | FF++ | | | CelebDFv2 | | | UADFV | | |
|---|---|---|---|---|---|---|---|---|---|
| Metric | ACC | EER | AUC | ACC | EER | AUC | ACC | EER | AUC |
| FF++ | 0.89 | 0.23 | 0.83 | 0.66 | 0.45 | 0.59 | 0.50 | 0.50 | 0.50 |
| CelebDFv2 | 0.50 | 0.57 | 0.49 | 0.59 | 0.58 | 0.48 | 0.40 | 0.71 | 0.33 |
| UADFV | 0.50 | 0.50 | 0.50 | 0.33 | 0.62 | 0.33 | 0.49 | 0.55 | 0.48 |
| Deepfakeface | 0.47 | 0.44 | 0.59 | 0.23 | 0.77 | 0.34 | 0.37 | 0.28 | 0.78 |
| DFDB-Face | 0.72 | 0.49 | 0.52 | 0.71 | 0.75 | 0.20 | 0.47 | 0.71 | 0.25 |
| JDB-face | 0.42 | 0.43 | 0.55 | 0.47 | 0.65 | 0.35 | 0.29 | 0.61 | 0.35 |
| Vanilla | 0.85 | 0.40 | 0.67 | 0.75 | 0.71 | 0.31 | 0.50 | 0.53 | 0.40 |
| KD | 0.84 | 0.37 | 0.71 | 0.81 | 0.35 | 0.65 | 0.50 | 0.59 | 0.48 |
| DW | 0.78 | 0.35 | 0.72 | 0.53 | 0.35 | 0.72 | 0.50 | 0.51 | 0.48 |
| MDB (ours) | **0.95** | **0.10** | **0.94** | **0.82** | **0.23** | **0.81** | **0.50** | **0.48** | **0.50** |

(b) Diffusion deepfake datasets (Deepfakeface, DFDB-Face, JDB-Face)

| Train Strategy | Deepfakeface | | | DFDB-Face | | | JDB-Face | | |
|---|---|---|---|---|---|---|---|---|---|
| Metric | ACC | EER | AUC | ACC | EER | AUC | ACC | EER | AUC |
| FF++ | 0.35 | 0.78 | 0.27 | 0.67 | 0.73 | 0.29 | 0.48 | 0.61 | 0.35 |
| CelebDFv2 | 0.25 | 0.80 | 0.17 | 0.49 | 0.83 | 0.20 | 0.23 | 0.82 | 0.20 |
| UADFV | 0.42 | 0.48 | 0.57 | 0.26 | 0.77 | 0.24 | 0.49 | 0.71 | 0.27 |
| Deepfakeface | 0.68 | 0.33 | 0.57 | 0.23 | 0.44 | 0.58 | 0.51 | 0.52 | 0.51 |
| DFDB-Face | 0.73 | 0.65 | 0.33 | 0.73 | 0.41 | 0.58 | 0.57 | 0.55 | 0.48 |
| JDB-face | 0.25 | 0.67 | 0.32 | 0.47 | 0.69 | 0.32 | 0.67 | 0.44 | 0.58 |
| Vanilla | 0.38 | 0.76 | 0.32 | 0.43 | 0.55 | 0.37 | 0.51 | 0.64 | 0.38 |
| KD | 0.72 | 0.34 | 0.68 | 0.76 | 0.32 | 0.67 | 0.57 | 0.62 | 0.40 |
| DW | 0.57 | 0.55 | 0.48 | 0.63 | 0.30 | 0.72 | 0.58 | 0.42 | 0.61 |
| MDB (ours) | **0.79** | **0.20** | **0.78** | **0.98** | **0.07** | **0.94** | **0.98** | **0.07** | **0.93** |

## 6 Conclusion

Diffusion models presents substantial challenges for real-world deepfake detection. This work addresses this urgency by introducing extensive diffusion deepfake datasets and highlighting the limitations of existing detection methods.

Our dataset is not only challenging to detect but is highly diverse compared to the present face deepfake datasets. We emphasize the crucial role of enhancing training data diversity on generalizability. Our proposed momentum difficulty boosting strategy, effectively tackles the challenge posed by training data heterogeneity. Extensive experiments show that our approach achieves state-of-the-art performance, surpassing prior alternatives significantly. It has shown high testing accuracy on the totally unknown dataset proving its generalizing ability. This work not only identifies the challenges of diffusion models in deepfake detection but also provides practical solutions, paving the way for more robust and adaptable countermeasures against the evolving threat of latest deepfakes.

# Supplementary Material

## 1 Dataset Construction Workflow

In this section, we have provided the visualization of the detailed workflow, complemented by a comprehensive visual representation of both misclassified and correctly classified samples encountered throughout the process.

### 1.1 JourneyDB-Face Dataset

Figures 3 through 6 showcase visual examples of both correctly classified and misclassified samples encountered during the process. Figures 3 and 4 illustrate the results of the metadata classification using BERT and the word filtering process, focusing respectively on the "Prompt" and "Style" sections. In the "Style" section, particular attention was given to filtering out images with an "anime style". Despite this filtering process, as depicted in these figures, the outcomes did not always align with expectations, particularly in cases where "anime style" was not explicitly mentioned in the "prompts". Instances of such misclassifications, along with their corresponding images, are displayed in Figure 5. Figure 6 demonstrates the results post-face filtering process, successfully isolating the intended images.

### 1.2 DiffusionDB-Face Dataset

The creation of DiffusionDB-Face involved a bit different approach compared to JourneyDB-Face, adapted to fit the format of the source dataset. As detailed in the main paper, the initial step entailed classifying prompts likely to generate images of human faces using BERT. For instance, Figure 7 displays BERT's classification scores for several samples, providing both "human face" and "not human face" evaluations. Despite this, some metadata were inaccurately classified due to specific words or structures in the "prompts", as illustrated in Figure 8. To mitigate this, face filtering was employed to exclude irrelevant images. However, as Figure 9 reveals, this method was not foolproof and occasionally included drawings or paintings of human faces. To address this, the Canny edge filter was applied to remove cartoon-styled images. In the main text (Section 3.1), we have addressed the detailed description of the Canny edge detector's threshold. This resulted in more precise outcomes, with some examples of the refined images presented in Figure 10.

## 2 Ablative Analysis

**Sensitivity Analysis of $C$:** (1) We examine the effect of the scale factor, $C$ (Eq 3) in the main text. As observed from Table 1, this parameter is not sensitive with a good range of selections.

Table 1: Ablation of the scale factor with MDB (Accuracy).

| $C$ | FF++ | CDFv2 | UADFV | DFF | DFDB | JDB |
|---|---|---|---|---|---|---|
| 1 | 0.87 | 0.78 | 0.48 | 0.69 | 0.73 | 0.74 |
| 3 | 0.91 | 0.78 | 0.49 | 0.72 | 0.74 | 0.74 |
| 5 | 0.95 | 0.82 | 0.50 | 0.79 | 0.98 | 0.98 |
| 7 | 0.94 | 0.79 | 0.51 | 0.75 | 0.81 | 0.79 |
| 9 | 0.92 | 0.79 | 0.49 | 0.75 | 0.81 | 0.81 |
| 10 | 0.91 | 0.79 | 0.50 | 0.73 | 0.81 | 0.80 |

**Frequency analysis:** We make visual analysis of frequency distributions across all datasets similar to Zhang *et al.* [59]. This elucidates the distinguishing characteristics between authentic and deepfake imagery. Figure 1 indicates that, the frequency distinction between authentic and synthetic images produced by diffusion models is generally more subtle and thus presents more challenges, compared to that in traditional datasets.

**Sample weight dynamics over training:** Figure 2 presents the per-dataset histogram of weights across training epochs with our MDB.

We note that DiffusionDB-Face and JourneyDB-Face datasets are assigned with highest weights, indicating more challenges presented. This difficulty aware training can benefit the performance (see Table 5 in main text).

## 3 Limitations

Despite the extensive filtering processes applied to the two substantial datasets, JourneyDB and DiffusionDB, there might remain a handful of instances where the images are either overly cartoonized or lack sufficient realism. These anomalies may be overlooked in subsequent stages, such as the further Face Filtering and the custom model designed for animated or human facial images. As highlighted in the primary paper concerning dataset statistics, there is a significant gender distribution disparity, originating from the source databases (likely due to the processes of both prompting and training the generative models).

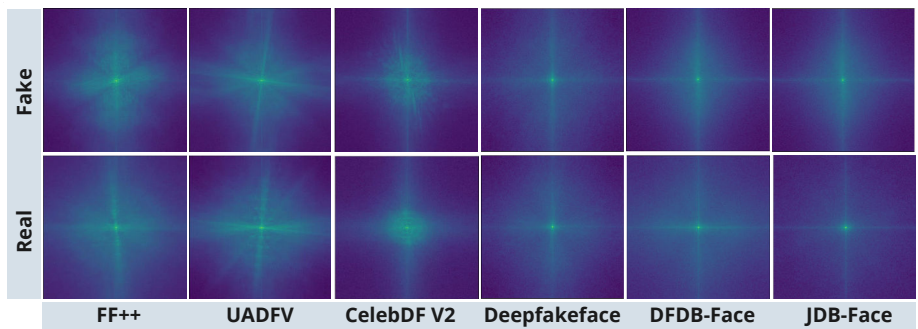

Fig. 1: Frequency analysis: the average spectra of each high-pass filtered image

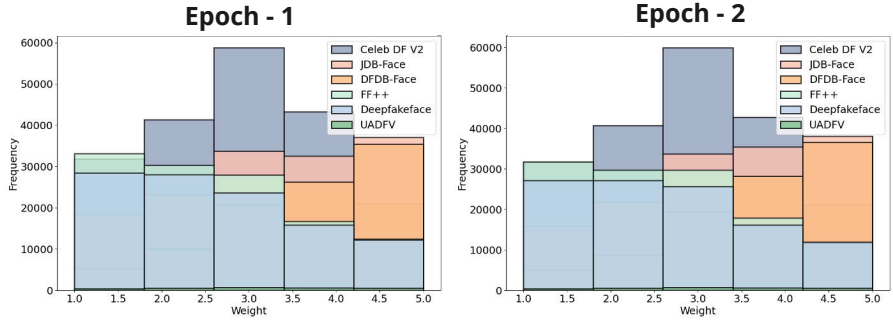

Fig. 2: Ablative Analysis: (a) Frequency analysis and (b) weight distribution and dynamics .

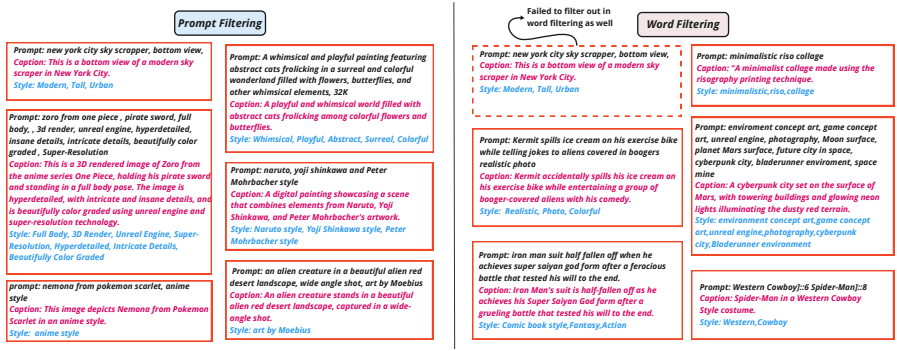

Fig. 3: JourneyDB-Face: Examples of misclassified metadata by BERT.

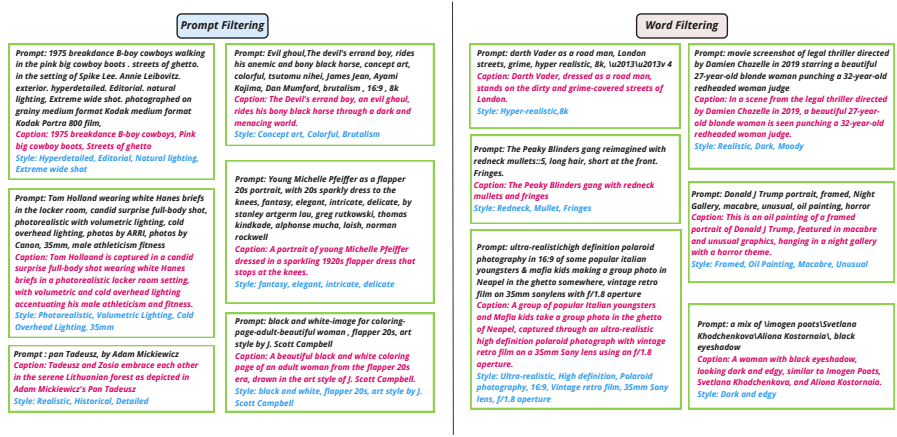

Fig. 4: JourneyDB-Face: Examples of correctly classified metadata by BERT.

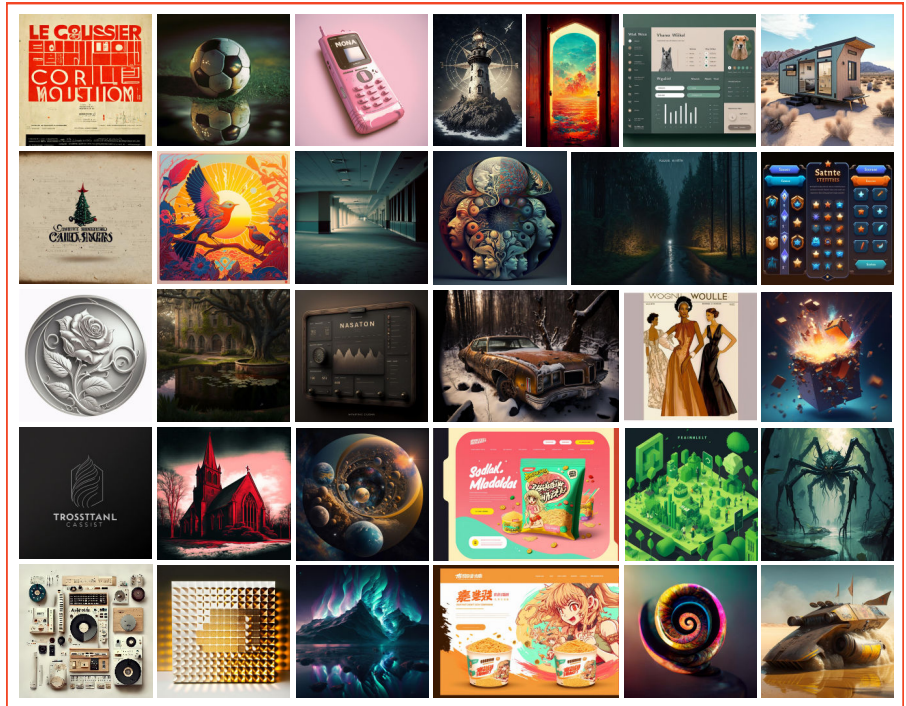

Fig. 5: JourneyDB-Face: Unfiltered samples in word filtering due to the absence of "Anime Style" mention.

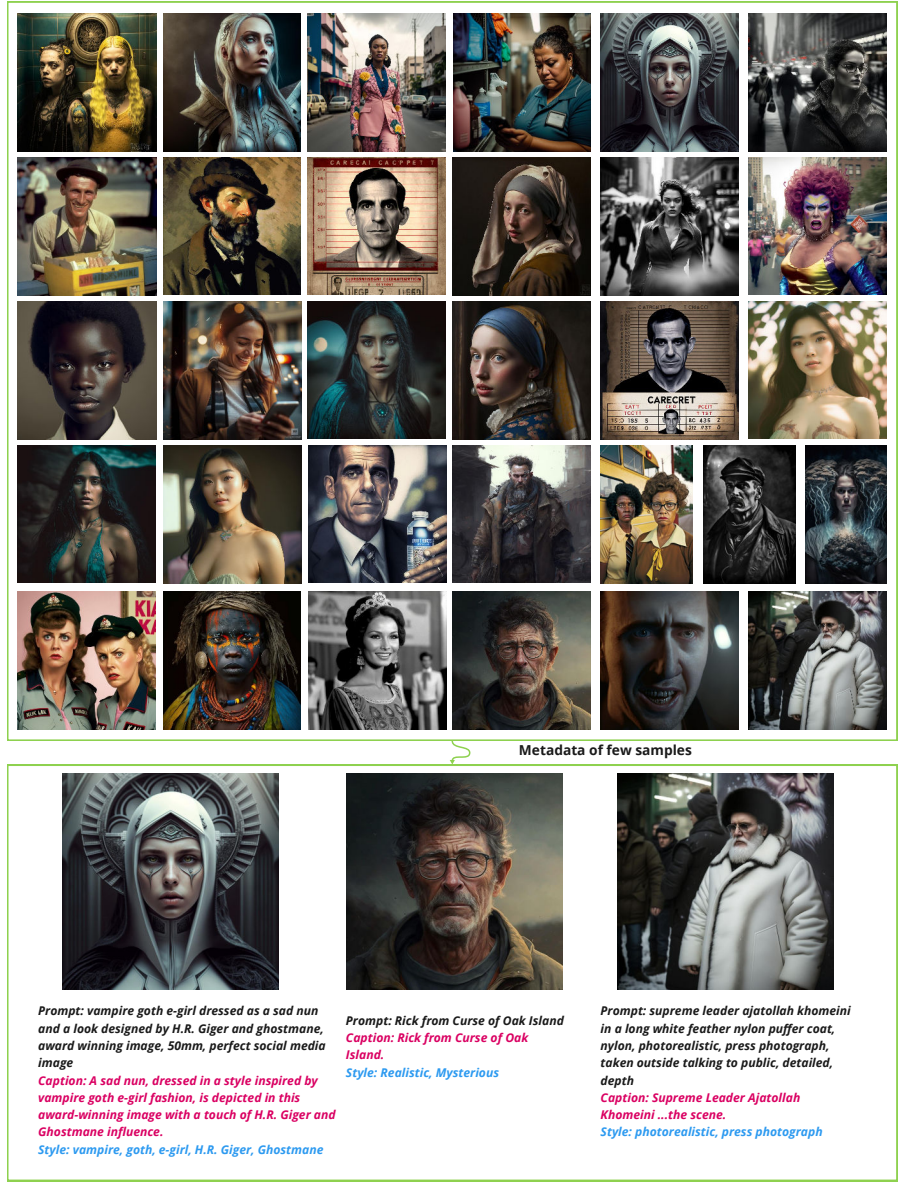

Fig. 6: JourneyDB-Face: Examples of correctly filtered samples after word filtering process.

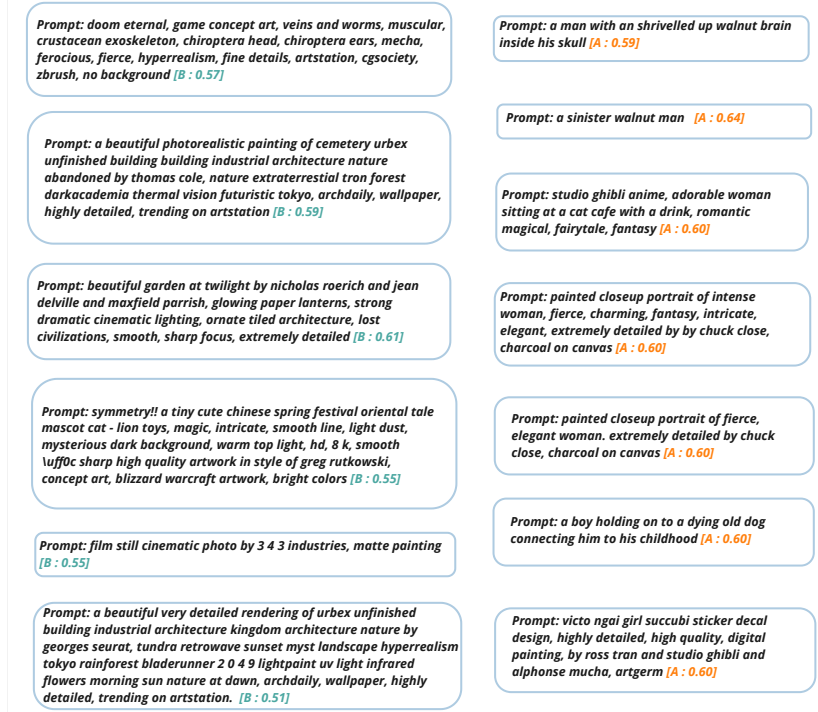

Fig. 7: DiffusionDB-Face : Examples of BERT classified metadata with the corresponding scores. A: "human face" ; B: "not human face".

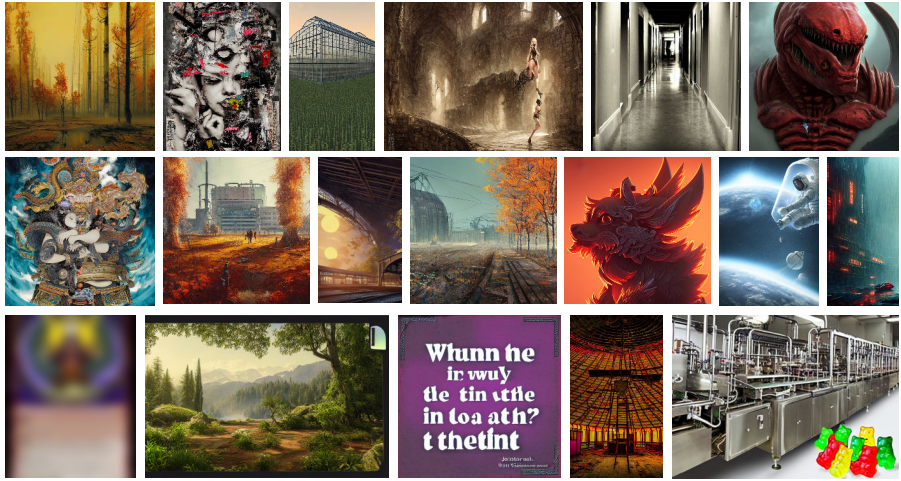

Fig. 8: Misclassified samples by BERT's metadata classification round.

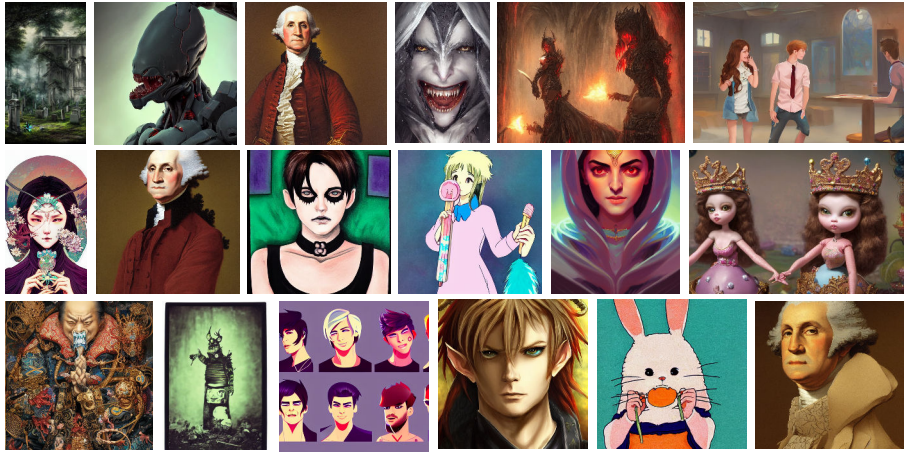

Fig. 9: DiffusionDB-Face: Animated face image samples after face filtering.

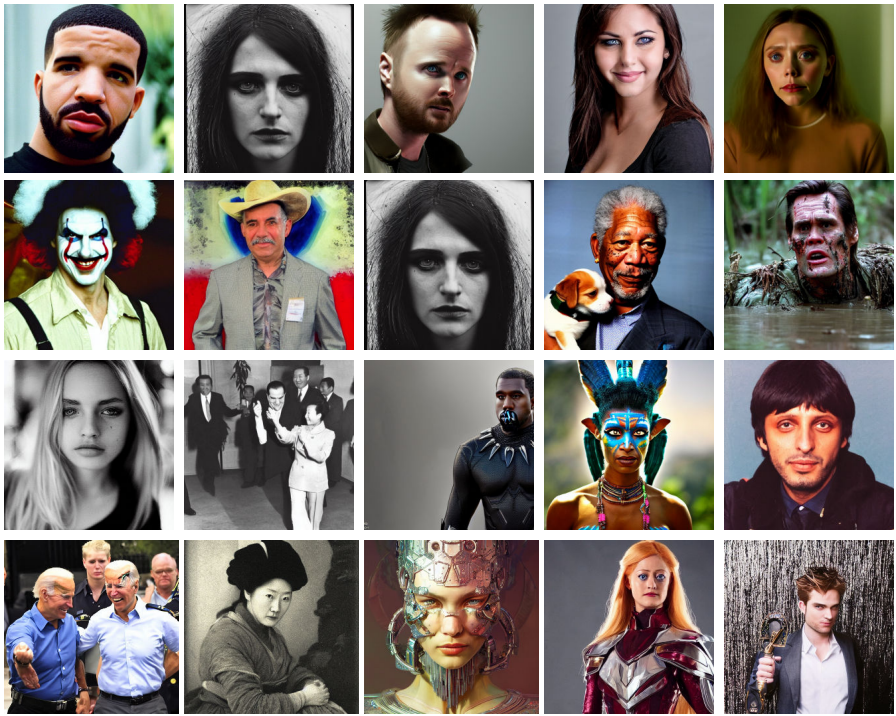

Fig. 10: DiffusionDB-Face: Few samples after applying Canny edge detector.